\documentclass{svproc}
\usepackage{url}
\usepackage{booktabs}
\usepackage{multirow}
\usepackage{amssymb,amsfonts}
\usepackage{amsmath,graphicx}
\usepackage[ruled,vlined]{algorithm2e}
\usepackage{caption}
\usepackage[]{hyperref}
\usepackage{array}

\usepackage{fancyhdr}
\fancypagestyle{specialfooter}{%
  \fancyhf{}
  
  \fancyfoot[L]{\scriptsize In the 10th International Conference on Complex Networks and Their Applications (2021), Madrid, Spain}
}

\newcommand{\xhdr}[1]{\vspace{1.3mm}\noindent{{\bf #1.}}}
\newcommand{\ie}{\textit{i.e.}}

\newsavebox{\leftbox}
\newsavebox{\rightbox}

\begin{document}

\mainmatter              
\title{Propagation on Multi-relational Graphs for Node Regression}
\titlerunning{Multi-relational Propagation}  
%
\author{Eda Bayram 
}
\authorrunning{Eda Bayram} 
%
\tocauthor{Eda Bayram}
\institute{\'{E}cole Polytechnique F\'{e}d\'{e}rale de Lausanne (EPFL), Lausanne, Switzerland\\
\email{eda.bayram@epfl.ch}}

\maketitle              

\begin{abstract}
Recent years have witnessed a rise in real-world data captured with rich structural information that can be conveniently depicted by multi-relational graphs. While inference of continuous node features across a simple graph is rather under-studied by the current relational learning research, we go one step further and focus on node regression problem on multi-relational graphs. We take inspiration from the well-known label propagation algorithm aiming at completing categorical features across a simple graph and propose a novel propagation framework for completing missing continuous features at the nodes of a multi-relational and directed graph. Our multi-relational propagation algorithm is composed of iterative neighborhood aggregations which originate from a relational local generative model. Our findings show the benefit of exploiting the multi-relational structure of the data in several node regression scenarios in different settings.
\end{abstract}
\thispagestyle{specialfooter}%
\section{Introduction}
Various disciplines are now able to capture different level of interactions between entities of their interest, which promotes multiple types of relationships within data. Examples include social networks \cite{cozzo2016multilayer,wasserman1994social}, biological networks \cite{bentley2016multilayer,de2017multilayer}, transportation networks \cite{boccaletti2014structure,aleta2019multilayer}, etc.
Multi-relational graphs are convenient for representing such complex network-structured data.
Recent years have witnessed a strong line of relational learning studies focusing on the inference of node-level and graph-level categorical features \cite{chami2020machine}. Most of these are working on simple graphs and there has been little interest in the regression of continuous node features across the graph. In particular, node regression on multi-relational graphs still remains unexplored.

In this study, we present a multi-relational node regression framework.
Given multi-relational structure of data and partially observed continuous features belonging to the data entities, we aim at completing missing features.
It is possible to encode intrinsic structure of the data by
a graph accommodating multiple types of directed edges between graph's nodes that represent the data entities.
Accordingly, we establish the main research question we address: \textit{How can we achieve node-value imputation on a multi-relational and directed graph?} For this purpose, we propose an algorithm which propagates observed set of node features towards missing ones across a multi-relational and directed graph. We take inspiration from the well-known label propagation algorithm \cite{zhu2002learning} aiming at completing categorical features across a simple, weighted graph.
We see that simple neighborhood aggregations operated on a given relational structure hold the basis for many iterative graph algorithms including the label propagation.
Thus, we first break down the propagation framework by the neighborhood aggregations derived through a simple local generative model. Later, we extend this by incorporating a multi-relational neighborhood and suggest a relational local generative model. Then, we build our algorithm, which we call multi-relational propagation (\textsc{MrP}), by iterative neighborhood aggregation steps originating from this new model.
We provide the derivation of the parameters of the proposed model, which can be estimated over the observed set of node features.
Our method can be considered as a sophisticated version of the standard propagation algorithm by enabling regression of continuous node features over a multi-relational and directed graph.
We compare our multi-relational propagation method against the standard propagation in several node regression scenarios. At each, our approach enhances the results considerably by integrating multi-relational structure of data into the regression framework.
\begin{figure}[t]
  \centering
 \includegraphics[width=0.5\textwidth]{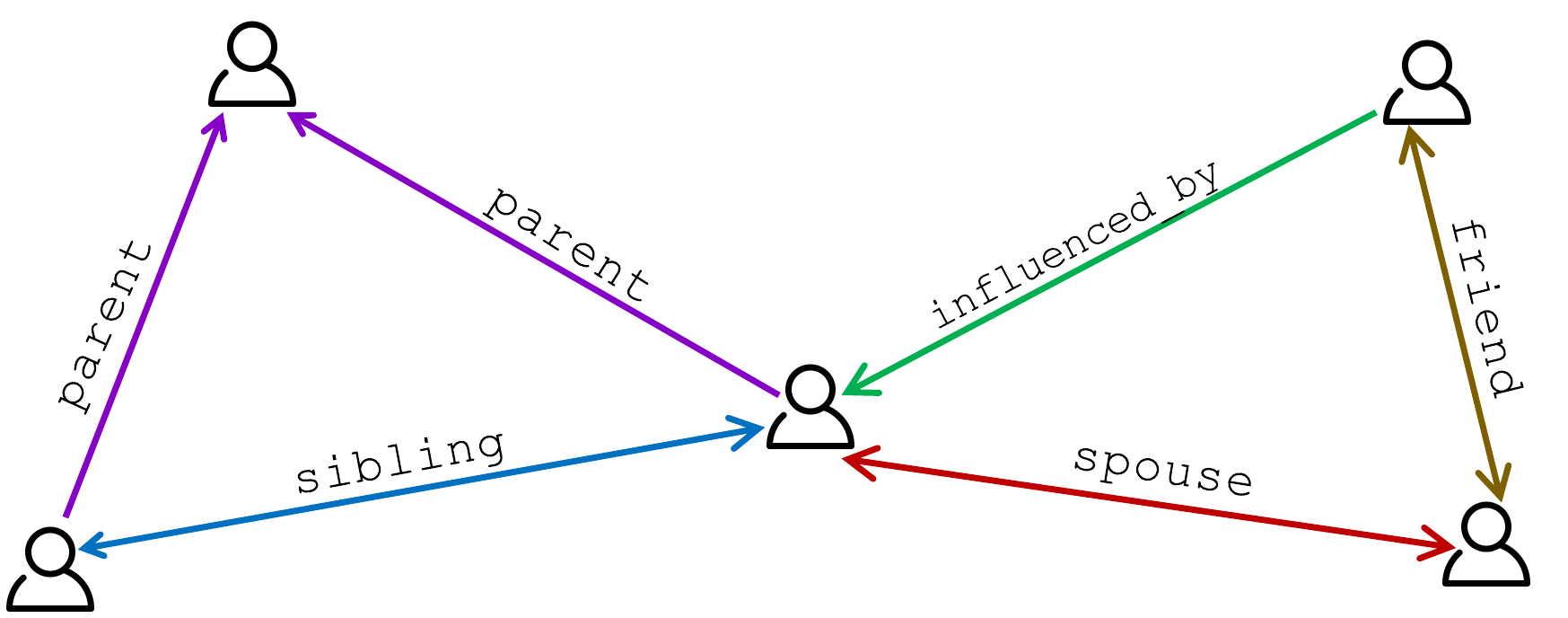}
 \vspace{-2mm}
	\caption{A fragment of a multi-relational and directed social network}
  \label{fig: illust}
\end{figure}

\xhdr{Comparison to existing schemes}
The node regression problem has been studied on simple graphs for signal inpainting \cite{chen2014signal,perraudin2017stationary} and node representation learning \cite{opolka2019spatio,wu2020graph,deng2021edge,ivanov2021boost}.
Many of these approaches implicitly employ a smoothness prior which promotes similar representations at the neighboring nodes of the graph \cite{zhou2004learning}.
The smoothness prior exploited in node representation learning studies broadly prescribes minimizing the Euclidean distance between features at the connected nodes. Throughout the paper, we refer to such prior as $\ell_2$ sense smoothness.
Despite its practicality, $\ell_2$ sense smoothness prior suffers from several major limitations that might mislead regression on a multi-relational and directed graph. First, it treats all neighbors of a node equally while reasoning about the node's state even though neighbors connected via different types of relations might play different roles in the inference task. For instance, Fig. \ref{fig: illust} illustrates multiple types of relationships that might arise between people. Here, each relation type presumably relies on a different affinity rule or different level of importance depending on the node regression task.
Second, some relation types are inherently symmetric, and some others are asymmetric\footnote{In directed graphs, symmetric relationships emerge from bi-directed edges where
the edge direction is valid in both directions such as \texttt{sibling} whereas in asymmetric relationships, the edge direction is valid in only one direction such as \texttt{parent}, \texttt{child}. See the edge directions in Fig \ref{fig: illust}.}.
Euclidean distance minimization broadly assumes that values at neighboring nodes are as close as possible, which may not always be the case for asymmetric relationships.
We thus depart from the straightforward $\ell_2$ sense of smoothness and augment the prior with a relational local generative model.

\xhdr{Contributions} In this study,
(i) we provide a breakdown of the propagation algorithm on simple graphs from the Bayesian perspective, (ii) we introduce a relational local generative model, which permits neighborhood aggregation operation on a multi-relational, directed neighborhood, (iii) we propose a novel propagation framework \textsc{MrP}, which properly handles propagating observed continuous node features across a multi-relational directed graph and complete missing ones.
\vspace{-2mm}
\section{Propagation on Simple Graphs}
\vspace{-2mm}
We denote a simple, undirected graph by  $\mathcal{G}(\mathcal{V},\mathcal{E})$ with set of nodes $\mathcal{V}$ and set of edges $\mathcal{E}$. Also, we denote $x_i \in \mathbb{R}$ as the continuous node feature\footnote{Generalization to vectorial node representations is possible in principle, yet omitted here for the sake of simplicity.} held by node-$i$.

\xhdr{Local generative model}
We recall the smoothness prior prescribing the neighboring node representations to be as close as possible in terms of $\ell_2$-norm. Consequently, we write a simple local generative model which relates two neighboring nodes as $x_i = x_j + \epsilon$
where $(i,j) \in \mathcal{E}$ and $\epsilon \sim \mathcal{N}(0, \sigma^2_{ij})$.

\xhdr{First-order Bayesian estimate of node's value} The local generative model can be used to obtain an approximation of the node's state in terms of its local neighborhood. This can be achieved by maximizing the expectation of the node's feature given that of its first-hop neighbors:
\begin{align}
\label{eqn: local Bayes}
\operatorname*{argmax}_{x_i} \quad
\mathrm{p}(x_i| \{x_j : (i,j) \in \mathcal{E}\}) =
\operatorname*{argmax}_{x_i} \quad
\frac{\mathrm{p}(\{x_j : (i,j) \in \mathcal{E}\}|x_i)\mathrm{p}(x_i)}
{\mathrm{p}(\{x_j : (i,j) \in \mathcal{E}\})},
\end{align}
where Bayes' rule applies.
Here, we make two assumptions. First, we assume that the prior distribution on the node features, $\mathrm{p}(x_i) \: \forall i \in \mathcal{V}$, is uniform. Second, we only consider the partial correlations between the central node---whose state is to be estimated---and its first-hop neighbors while we neglect any partial correlation among the neighbor set---conditionally independence assumption. Accordingly, we reformulate the problem as
\begin{align}
\label{eqn: MLE}
\operatorname*{argmax}_{x_i} \:
\prod_{(i,j) \in \mathcal{E}}\mathrm{p}(x_j|x_i) \quad =
\operatorname*{argmin}_{x_i} \:
- \sum_{(i,j) \in \mathcal{E}}\mathrm{log}(\mathrm{p}(x_j|x_i)),
\end{align}
and rewrite it as minimization of negative log-likelihood. Next, we plug in the local generative model and obtain the following problem:
\begin{align}
\label{eqn: minim}
\operatorname*{argmin}_{x_i} \quad
\sum_{(i,j) \in \mathcal{E}}\frac{\|x_j - x_i\|_2^2}{\sigma_{ij}^2}.
\end{align}

\xhdr{Neighborhood aggregation}
The first-order Bayesian estimate boils down to minimizing the Euclidean distance between node's feature to that of its neighbors, \ie, suggesting a least squares problem in \eqref{eqn: minim}. Its solution is simply found by setting the gradient of the objective to zero:
\begin{align}
\label{eqn: Bayes est}
\hat{x}_i =
\frac{\sum_{(i,j) \in \mathcal{E}} \omega_{ij} x_j}
{\sum_{(i,j) \in \mathcal{E}} \omega_{ij}},
\end{align}
where $\omega_{ij} = 1/\sigma_{ij}^2$. As seen, it is a linear combination of the neighbors' features.
\begin{figure}[t]
  \centering
 \includegraphics[width=1\textwidth]{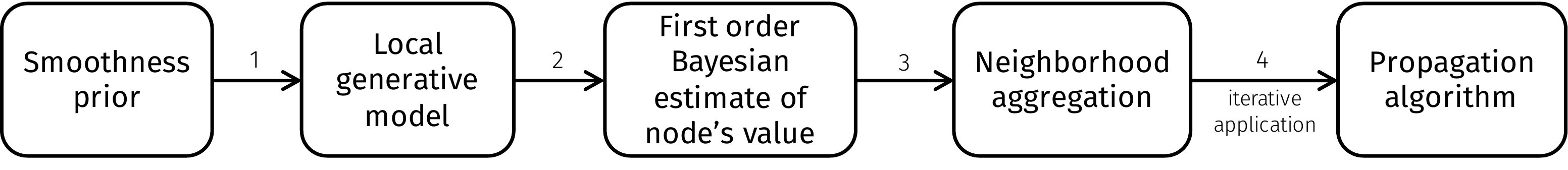}
	\caption{Overview of the pipeline for the development of a propagation algorithm}
  \label{fig: pipeline}
\end{figure}
The first-order Bayesian estimation in the conditions considered above clarifies the neighborhood aggregation operation accomplished in one iteration of a propagation algorithm \cite{zhu2002learning}.
Estimating the node states across the whole graph iteratively, a propagation algorithm expands the scope of the approximation beyond the first-order until a stopping criterion is satisfied. Hence, we summarize the pipeline for developing a propagation algorithm as given in Fig. \ref{fig: pipeline}.
\vspace{-2mm}
\section{Multi-relational Model}
\vspace{-2mm}
We now introduce a multi-relational and directed graph as $\mathcal{G}(\mathcal{V},\mathcal{E}, \mathcal{P})$, where $\mathcal{V}$ is the set of nodes, $\mathcal{P}$ is the set of relation types, $\mathcal{E} \subseteq  \mathcal{V} \times  \mathcal{P} \times \mathcal{V}$ is the set of multi-relational edges. The function $\mathtt{r}(i,j)$ returns the relation type $\mathtt{p} \in \mathcal{P}$ that is pointed from node $j$ to node $i$. If such a relation exists between them, yet pointed from the node $i$ to the node $j$, the function returns the reverse as $\mathtt{p}^{-1}$.

\xhdr{Relational local generative model}
It is required to diversify the simple local generative model by the set of relationships existing on a multi-relational graph. To this end, we propose the following local generative model for the node's state given its multi-relational and directed neighbors:
\begin{equation}
\label{eqn: rel gen model}
    x_i =
    \Bigg\{
      \begin{array}{l}
        \eta_{\mathtt{p}} x_j + \tau_{\mathtt{p}} + \epsilon, \quad	 \forall \mathtt{r}(i,j) = \mathtt{p} \text{ where } \epsilon \sim \mathcal{N}(0,\sigma_{\mathtt{p}}^2)\\ 
        \cfrac{x_j}{\eta_{\mathtt{p}}}  - \cfrac{\tau_{\mathtt{p}}}{\eta_{\mathtt{p}}} + \epsilon, \quad	 \forall \mathtt{r}(i,j) = \mathtt{p}^{-1} \text{ where } \epsilon \sim \mathcal{N}(0,\cfrac{\sigma_{\mathtt{p}}^2}{\eta_{\mathtt{p}}^2}).
      \end{array}
\end{equation}
This model builds a linear relationship between neighboring nodes by introducing relation-dependent scaling parameter $\eta$ and a shift parameter $\tau$. The latter case in \eqref{eqn: rel gen model} indicates the generative model yielded by the reverse relation, where the direction of the edge is reversed. Such a linear model conforms both symmetric and asymmetric relationships. This is because it can capture any bias over a certain relation through parameter $\tau$ or any change in scale through parameter $\eta$. We note that the default set for these parameters are suggested as $\tau = 0, \eta = 1$, which boils down to the simple local generative model.

\xhdr{First-order Relational Bayesian Estimate}
We now estimate the node's state by its first-hop neighbors connected via multiple types relationships. We repeat the same assumptions as in \eqref{eqn: local Bayes}, which casts the problem as maximizing the likelihood of node's first-hop neighbors.
Once the likelihood of relational neighbors is expressed through the model in \eqref{eqn: rel gen model}, the estimation can be found by minimizing the negative log-likelihood as in \eqref{eqn: MLE}. Consequently, we obtain the following objective:
\vspace{-2mm}
{\small
\begin{align}
\label{eqn: rel Bayes problem}
\operatorname*{argmin}_{x_i} \:
\sum_{\mathtt{p} \in \mathcal{P}}\Bigg(
\sum_{\mathtt{r}(i,j) = \mathtt{p}}
\cfrac{\omega_\mathtt{p}}{2}\Big(x_i - \eta_\mathtt{p}x_j -\tau_\mathtt{p}\Big)^2
+\sum_{\mathtt{r}(i,j) = \mathtt{p}^{-1}}
\cfrac{\omega_\mathtt{p}\eta_\mathtt{p}^2}{2}\Big(x_i - \cfrac{x_j}{\eta_\mathtt{p}} +\cfrac{\tau_\mathtt{p}}{\eta_\mathtt{p}}\Big)^2
\Bigg),
\end{align}}
where we apply a change of parameter $\omega_p =1/\sigma^2_{\mathtt{p}}$.

\vspace{2mm}
\xhdr{Relational Neighborhood Aggregation}
For an arbitrary node $i \in \mathcal{V}$, we denote the loss to be minimized as $\mathcal{L}_i$. Such a loss leads to a least squares problem whose solution satisfies $\cfrac{\partial\mathcal{L}_i}{\partial x_i}(\hat{x}_i) = 0$. Accordingly, the estimate can be found as
\vspace{-4mm}
{\small
\begin{align}
\label{eqn: rel estim}
    \hat{x}_i =
    \cfrac{
    \sum_{\mathtt{p} \in \mathcal{P}}\Big(
\sum_{\mathtt{r}(i,j) = \mathtt{p}} \:
\omega_\mathtt{p}\Big(\eta_\mathtt{p}x_j +\tau_\mathtt{p}\Big)+
\sum_{\mathtt{r}(i,j) = \mathtt{p}^{-1}} \:
\omega_\mathtt{p}\eta^2_\mathtt{p}\Big(\cfrac{x_j}{\eta_\mathtt{p}} -\cfrac{\tau_\mathtt{p}}{\eta_\mathtt{p}}\Big)\Big)}
    {\sum_{\mathtt{p} \in \mathcal{P}}\Big(
\sum_{\mathtt{r}(i,j) = \mathtt{p}}\:
\omega_\mathtt{p}+
\sum_{\mathtt{r}(i,j) = \mathtt{p}^{-1}}\:
\omega_\mathtt{p}\eta^2_\mathtt{p}\Big)}.
\end{align}}
\vspace{-8mm}
\subsection{Estimation of Relational Parameters}
\label{sec: estim param}
The parameters of the local generative model associated with relation type $\mathtt{p} \in \mathcal{P}$ are introduced as $\{\tau_\mathtt{p}, \eta_\mathtt{p}\, \omega_\mathtt{p}\}$. These parameters can be estimated over the set of node pairs connected to each other by relation $\mathtt{p}$, \ie, $\big\{(x_i,x_j) \: \forall i,j \in \mathcal{V}\: | \: \mathtt{r}(i,j)=\mathtt{p}\big\}$. For this purpose, we carry out the maximum likelihood estimation over the parameters:
\vspace{-4mm}
{\small
\begin{align}
    \operatorname*{argmax}_{\tau_\mathtt{p}, \eta_\mathtt{p}\, \omega_\mathtt{p}} \quad
    \mathrm{p}\Big(
    \big\{(x_i,x_j) \: \forall i,j \in \mathcal{V}\: | \: \mathtt{r}(i,j)=\mathtt{p} \big\} \: \big| \:
    \tau_\mathtt{p}, \eta_\mathtt{p}\, \omega_\mathtt{p} \Big).
\end{align}}
Then, we conduct an approximation over the node pairs that are connected by a given relation type while neglecting any conditional dependency that might exist among these node pairs\footnote{A first-order approximation is conducted where each node pair connected via a certain relation type is considered as an independent observation although these node pairs might appear in the same neighborhood.}. Hence, we can write the likelihood on each node pair in a product as follows:
\vspace{-2mm}
{\small
\begin{align}
\label{eqn: rel MLE}
    \operatorname*{argmax}_{\tau_\mathtt{p}, \eta_\mathtt{p}\, \omega_\mathtt{p}} \quad
    \prod_{\mathtt{r}(i,j)=\mathtt{p}}
    \mathrm{p}\Big(
    (x_i,x_j)\: \big| \: \tau_\mathtt{p}, \eta_\mathtt{p}\, \omega_\mathtt{p}
    \Big).
\end{align}}
We proceed with the minimization of negative log-likelihood to solve \eqref{eqn: rel MLE}. The reader might recognize that its solution is equivalent to the parameters of a linear regression model \cite{rencher2012}. This is simply because we introduce linear generative models \eqref{eqn: rel gen model} for the relationships existing on the graph. Therefore, the parameters can be found as follows ($\mu = \mathrm{mean}(\mathbf{x})$ is the mean of node values):
\begin{align}
\label{eqn:eta}
\resizebox{.91\hsize}{!}{$
\eta_{\mathtt{p}} = \cfrac{\sum_{\mathtt{r}(i,j)=\mathtt{p}}
(x_i- \mu)(x_j- \mu)}
{\sum_{\mathtt{r}(i,j)=\mathtt{p}}
(x_j- \mu)^2}, \:
\begin{array}{l}
\tau_{\mathtt{p}} = \mathrm{mean}\Big(
\big\{(x_i - \eta_{\mathtt{p}} x_j) 
\: \forall i,j \in \mathcal{V}\: | \: \mathtt{r}(i,j)=\mathtt{p} \big\}
\Big),\\
\omega_\mathtt{p} = 1/\mathrm{mean}\Big(
\big\{(x_i - \eta_\mathtt{p} x_j - \tau_{\mathtt{p}})^2
\: \forall i,j \in \mathcal{V}\: | \: \mathtt{r}(i,j)=\mathtt{p} \big\}
\Big).
\end{array}
$}
\end{align}

\xhdr{Local Generative Model and Local Operation}
We summarize the local generative model, the associated loss and the first order estimate in simple and multi-relational cases in Table \ref{tbl: recap}.
In the multi-relational case, the neighborhood aggregation is not directly a weighted average of the neighbors but the neighbors are subject to a transformation with respect to the type and the direction of their relation to the central node. The relational transformation is controlled by the parameters $\eta$ and $\tau$. For this reason, in Table \ref{tbl: recap} we use the following functions as shortcuts for the transformations applied on the neighbors: $f(x) = x$ in simple case---no actual transformation applied, and $f_\mathtt{p}(x) = \eta_\mathtt{p}x + \tau_\mathtt{p}$ in relational case for type $\mathtt{p}$.
In addition, $\mathcal{P}^{-1} = \{\mathtt{p}^{-1}, \forall \mathtt{p} \in \mathcal{P}\}$ denotes the set relation types where the edge direction is reversed. For the reversed relationships, the set of parameters can be simply set as
$\eta_{\mathtt{p}^{-1}} = 1/{\eta_{\mathtt{p}}}$,  
$\tau_{\mathtt{p}^{-1}} = \tau_{\mathtt{p}}/{\eta_{\mathtt{p}}}$, 
$\omega_{\mathtt{p}^{-1}} = \eta_\mathtt{p}^2\omega_\mathtt{p}$.
Subsequent to the transformations, the estimation is computed by a weighted average that is controlled by the parameter $\omega$. It is worth to notice that $\omega$ is equivalent to the inverse of the error variance of the relational local generative model \eqref{eqn: rel gen model}. Therefore, the estimate can be interpreted as the outcome of an aggregation with precision that ranks the relational information.
\vspace{-4mm}
\begin{table}[h]
\centering
\caption{Local Generative Model and Operation}
\label{tbl: recap}
\resizebox{1\textwidth}{!}{
\begin{tabular}{m{0.13\linewidth} | c | c| c}
\toprule
 & \textbf{Local Generative Model} & \textbf{Loss} & \textbf{Local Operation}\\
\midrule
\textbf{Simple \newline Weighted \newline Graph} &
$\begin{array} {l} x_i = x_j + \epsilon \\ \forall(i,j) \in \mathcal{E} \\
\epsilon \sim \mathcal{N}(0,1/\omega_{ij})\end{array}$ &
$\begin{array} {c} \sum\limits_{(i,j)\in \mathcal{E}} \omega_{ij}(x_i-x_j)^2 \end{array}$&
$\begin{array} {l} \cfrac{\sum\limits_{(i,j)\in \mathcal{E}} \omega_{ij}f(x_j)}{\sum\limits_{(i,j)\in \mathcal{E}} \omega_{ij}} \end{array}$ \\
\hline
\textbf{Multi-relational \newline Directed \newline Graph} & 
$\begin{array} {l} x_i = \eta_\mathtt{p} x_j + \tau_\mathtt{p} + \epsilon \\ \forall \mathtt{r}(i,j) = \mathtt{p} \\
\epsilon \sim \mathcal{N}(0,1/\omega_\mathtt{p})\end{array}$ &
$\begin{array} {l} \sum\limits_{\mathtt{p}\in \mathcal{P}\cup \mathcal{P}^{-1}}
\sum\limits_{\mathtt{r}(i,j) = \mathtt{p}} \omega_\mathtt{p}(x_i-\eta_\mathtt{p}x_j - \tau_\mathtt{p})^2 \end{array}$ &
$\begin{array} {c} \cfrac{\sum\limits_{\mathtt{p}\in \mathcal{P}\cup \mathcal{P}^{-1}}\sum\limits_{\mathtt{r}(i,j) = \mathtt{p}} \omega_\mathtt{p}f_\mathtt{p}(x_j)}
{\sum\limits_{\mathtt{p}\in \mathcal{P}\cup \mathcal{P}^{-1}}\sum\limits_{\mathtt{r}(i,j) = \mathtt{p}} \omega_\mathtt{p}} \end{array}$\\
\bottomrule
\end{tabular}}
\end{table}
\vspace{-7mm}
\subsection{Multi-relational Propagation Algorithm}
\vspace{-1mm}
In Fig. \ref{fig: pipeline}, the propagation algorithm is depicted as an iterative neighborhood aggregation method where each iteration computes the solution of a first-order Bayesian estimation problem.
Similarly, we propose a propagation algorithm that relies on the first-order relational Bayesian estimate that is introduced in \eqref{eqn: rel Bayes problem}. The algorithm operates iteratively where the relational neighborhood aggregation \eqref{eqn: rel estim} is accomplished at each node of the graph simultaneously. Thus, we denote a vector $\mathbf{x}^{(k)} \in \mathbb{R}^N$ composing the values at iteration-$k$ over the set of nodes for $|\mathcal{V}| = N$. Next, we express the iterations in matrix-vector multiplication format.

\xhdr{Iterations in matrix notation} We denote matrix $\mathbf{A}_\mathtt{p}$ for encoding the adjacency pattern of relation type $\mathtt{p}$. It is $(N \times N)$ asymmetric matrix storing incoming edges on its rows and outgoing edges on its columns.
One can compile aggregations in \eqref{eqn: rel estim} simultaneously over the entire graph using a matrix notation. Then, the relational local operations at iteration-$k$ can be expressed as
{\scriptsize
\begin{multline}
    \label{eqn: iter}
\mathbf{x}^{(k)} = 
\Bigg( \sum_{\mathtt{p}\in \mathcal{P}} \bigg( \omega_\mathtt{p}\big(\eta_\mathtt{p}\mathbf{A}_\mathtt{p}\mathbf{x}^{(k-1)}+ \tau_\mathtt{p}\mathbf{A}_\mathtt{p}\mathbf{1}\big)+
\omega_\mathtt{p}\eta_\mathtt{p}
\big(\mathbf{A}_\mathtt{p}^\top \mathbf{x}^{(k-1)}- \tau_\mathtt{p}\mathbf{A}_\mathtt{p}^\top\mathbf{1}\big) \bigg)
\Bigg)\\ \odot 
\Bigg(\sum_{\mathtt{p}\in \mathcal{P}} \bigg( \omega_\mathtt{p}\mathbf{A}_\mathtt{p}\mathbf{1} + \omega_\mathtt{p}\eta_\mathtt{p}^2\mathbf{A}_\mathtt{p}^\top\mathbf{1}
\bigg) \Bigg)^{-1},
\end{multline}}
where $\mathbf{1}$ is the vector of ones, $\odot$ stands for element-wise multiplication. In addition, the inversion on the latter sum term is applied element-wise. This part, in particular, arranges the denominator in Eqn. \eqref{eqn: rel estim} in vector format. Thus, it can be seen as the normalization factor over the neighborhood aggregation. For the purpose of simplification, we re-write \eqref{eqn: iter} as
\vspace{-2mm}
{\small
\begin{align}
    \mathbf{x}^{(k)} =
    (\mathbf{T}\mathbf{x}^{(k-1)} + \mathbf{S}\mathbf{1})
    \odot (\mathbf{H}\mathbf{1})^{-1},
\end{align}}
\vspace{-2mm}
by introducing the auxiliary matrices
{\small
\begin{equation}
\mathbf{T} = \sum_{\mathtt{p} \in \mathcal{P}} \eta_\mathtt{p}\omega_\mathtt{p} (\mathbf{A}_\mathtt{p} + \mathbf{A}_\mathtt{p}^\top),\:
\mathbf{S} = \sum_{\mathtt{p} \in \mathcal{P}} \tau_\mathtt{p}\omega_\mathtt{p} (\mathbf{A}_\mathtt{p} - \eta_\mathtt{p} \mathbf{A}_\mathtt{p}^\top),\:
\mathbf{H} = \sum_{\mathtt{p} \in \mathcal{P}} \omega_p(\mathbf{A}_\mathtt{p} + \eta_\mathtt{p}^2 \mathbf{A}_\mathtt{p}^\top).
\end{equation}}

\vspace{-2mm}
\xhdr{Algorithm} Given the iterations above, we now formalize the Multi-relational Propagation algorithm (\textsc{MrP}). \textsc{MrP} targets a node-level completion task where the multi-relational graph $\mathcal{G}$ is a priori given and the nodes are partially labeled at $\mathcal{U} \subseteq  \mathcal{V}$.
To manage propagation of continuous values at the labeled set of nodes towards the unlabeled, we introduce an indicator vector $\mathbf{u} \in \mathbb{R}^N$, which encodes the labeled nodes. It is initialized as
$\mathbf{u}^{(0)}_i = 1, \text{ if } i \in \mathcal{U}, \text{ else } 0$. Then, the vector $\mathbf{x}$ stores the node values throughout the iterations. It is initialized with the values over $\mathcal{U}$, and zero-padded at the unlabeled nodes, i.e., $\mathbf{x}^{(0)}_i = 0 \text{ if } i \in \mathcal{V} \setminus \mathcal{U}$.

Similar to the label propagation \cite{zhu2002learning}, our algorithm fundamentally consists of aggregation and normalization steps. In order to encompass multi-relational transformations during aggregation, we formulate an iteration of \textsc{MrP} by the steps of aggregation, shift and normalization. In addition, similar to the Page-rank algorithm \cite{brin1998anatomy}, we employ a damping factor $\xi \in [0,1]$ to update a node's state by combining its value from the previous iteration.
We provide a pseudocode for \textsc{MrP} in Algorithm \ref{alg: MrP}.
The propagation parameters for each relation type, $\{\tau_\mathtt{p}, \eta_\mathtt{p}, \omega_\mathtt{p}\}$ are estimated over the labeled set of nodes $\mathcal{U}$, as described in Section \ref{sec: estim param} and given to the algorithm as input together with the adjacency matrices encoding the multi-relational, directed graph.
Steps 1-4 in Algorithm \ref{alg: MrP} are essentially responsible for the multi-relational neighborhood aggregation. At Step-5, nodes' states are updated based on the collected information from the neighbors. If valid information collected from neighbors and the node is labeled, then we employ the damping factor, $\xi$, to update node's state. This adjusts the amount of trade-off between the neighborhood aggregation and the previous state of the node.
We distinguish whether an arbitrary node is currently labeled or not by the indicator vector, $\mathbf{u}^{(k)}$, which keeps track of propagated nodes throughout the iterations and
ensures that the normalization complies with valid collected neighborhood information. Hence, at Step 6, we update it as well. Finally, at Step 7, we clamp labeled set of nodes\footnote{Clamping step also exists in label propagation algorithm \cite{zhu2002learning}, which provides re-injection of true labels at each iteration throughout the propagation instead of overwriting the labeled nodes with the aggregated neighborhood information.} by leaving their values unchanged, simply because they store the governing information for completing the missing features. The algorithm terminates when all the nodes are propagated and the difference between two consecutive iterations is under a certain threshold. Accordingly, the number of iterations is related to the choice of hyperparameter $\xi$ and the stopping criterion.
\textsc{MrP}
is implemented using PyTorch-scatter package\footnote{Source code is available at https://github.com/bayrameda/MrAP.
A special case of \textsc{MrP} is studied to propagate heterogeneous node features in \cite{bayram2021node} for numerical attribute completion in knowledge graphs.},
which efficiently computes neighborhood aggregation on a sparse relational structure. Thus, the aggregation steps require $2|\mathcal{E}|$ operations, then, normalization and update steps require $|\mathcal{V}|$ operations at each iteration. Therefore, \textsc{MrP} scales linearly with the number of edges in the graph, similar to the standard label propagation algorithm (\textsc{LP}).
We finally note that by setting $\tau_\mathtt{p} = 0, \: \eta_\mathtt{p}=1,\: \omega_\mathtt{p}=1 \: \forall \mathtt{p} \in \mathcal{P}$, \textsc{MrP} drops down to \textsc{LP}\footnote{The label propagation algorithm \cite{zhu2002learning} was originally designed for completing categorical features across a simple, weighted graph. We render it to propagate continuous features and to be applicable for the node regression by the default parameter set of \textsc{MrP}.}
as if we operate on a simple graph regardless of the relation types and directions.
\vspace{-4mm}
\begin{algorithm}
{\scriptsize
\caption{\textsc{MrP}}
\label{alg: MrP}
 \textbf{Input:} $\mathcal{U}, \{x_i | i \in \mathcal{U}\}, \{\mathbf{A}_\mathtt{p}, \tau_\mathtt{p}, \eta_\mathtt{p}, \omega_\mathtt{p} \}_\mathcal{P}$\\
 \textbf{Output:} $\{x_i | i \in \mathcal{V} \setminus \mathcal{U}\}$\\
 \textbf{Initialization:} $\mathbf{u}^0, \mathbf{x}^0, \mathbf{T}, \mathbf{S}, \mathbf{H}$ \\
 \For{$k =1,2,\cdots$}{
 \textbf{Step 1. Aggregate:}
 $\mathbf{z} = \mathbf{T}\mathbf{x}^{(k-1)}$\\
 \textbf{Step 2. Shift:}
  $\mathbf{z} = \mathbf{z} + \mathbf{S}\mathbf{u}^{(k-1)}$\\
  \textbf{Step 3. Aggregate the normalization factors:}
  $\mathbf{r} = \mathbf{H}\mathbf{u}^{(k-1)}$\\
  \textbf{Step 4. Normalize:}
  $\mathbf{z} = \mathbf{z} \odot \mathbf{r}^\dagger$ \qquad \text{//$\dagger $ is for element-wise pseudo-inverse}\\
    \textbf{Step 5. Update values:}
  $$
  \mathbf{x}^{(k)}_i= \Bigg\{
\begin{array}{lll}
   \mathbf{x}^{(k-1)}_i, & \text{if } \mathbf{r}_i = 0  & \text{   // null info at neighbors}\\
   \mathbf{z}_i, & \text{if } \mathbf{r}_i > 0, \mathbf{u}^{(k-1)}_i = 0  & \text{   // null info at the node}\\
   (1- \xi)\mathbf{x}^{(k-1)}_i + \xi \mathbf{z}_i, &  \text{e.w.}
   (\mathbf{r}_i > 0, \mathbf{u}^{(k-1)}_i = 1)  & 
\end{array}
  $$ \\
  \textbf{Step 6. Update propagated nodes:}
  $ \mathbf{u}^{(k)} = \mathbf{u}^{(k-1)}, \quad \mathbf{u}^{(k)}_i = 1 \text{ if } \mathbf{r}_i > 0 $\\
  \textbf{Step 7. Clamp the known values:}
  $\mathbf{x}^{(k)}_i = x_i, \forall i \in \mathcal{U}$\\}
 \textbf{break} if
 $\mathrm{all}(\mathbf{u}^{(k)}) \: \& \: \mathrm{all}(\mathbf{x}^{(k)}-\mathbf{x}^{(k-1)} < \varepsilon)$\\
 $x_i = \mathbf{x}^{(k)}_i, \forall i \in \mathcal{X} \setminus \mathcal{U}$.}
\end{algorithm}

\vspace{-10mm}
\section{Experiments}
\vspace{-4mm}
We now present a proof of concept of the proposed multi-relational propagation method in several node regression scenarios in different settings. We first test \textsc{MrP} in estimating weather measurements on a multi-relational and directed graph that connects weather stations. Then, we evaluate the performance in predicting people's date of birth, where people are connected to each other on a social network composing different types relationships.

In the experiments, the damping factor is set as $\xi = 0.5$, then the threshold for terminating the iterations is fixed to $0.1 \%$ of the range of given values.
As evaluation metrics, we use root mean square error (RMSE), mean absolute percentage error (MAPE) and normalized RMSE (nRMSE) with respect to the range of groundtruth values. We calculate them over the estimation error on the unlabeled set of nodes. In the experiments, we leave the parameter $\eta$ in \textsc{MrP} by default as $1$ since we do not empirically observe a scale change over the relation types given by the datasets we work on. Then, we estimate the parameter $\tau$ and $\omega$ for each relation type based on the observed set of node values as described in Section \ref{sec: estim param}.
\vspace{-4mm}
\subsection{Multi-relational Estimation of Weather Measurements}
\vspace{-2mm}
We test our method on a meteorological dataset provided by MeteoSwiss, which compiles various types of weather measurements on 86 weather stations between years 1981-2010\footnote{https://github.com/bayrameda/MaskLearning/tree/master/MeteoSwiss}. In particular, we use yearly averages of weather measurements in our experiments.

\begin{figure}[t]
\centering
\begin{minipage}{0.8\textwidth}%
      \centering
        \includegraphics[width=\linewidth]{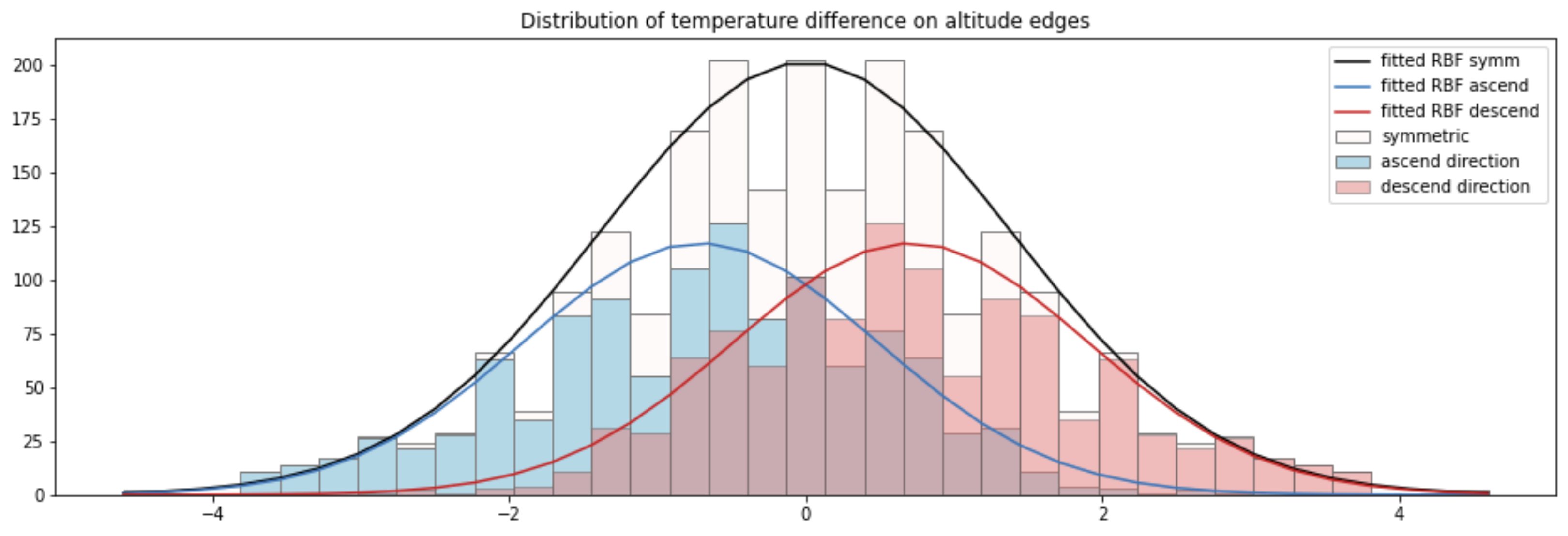}
    \end{minipage}
    \vspace{-2mm}
    \caption{Distribution of change in temperature measurements between weather stations that are related via altitude proximity in ascend and descend direction.}
\label{fig: Distribution alt}
\vspace{-6mm}
\end{figure}

\vspace{-2mm}
\xhdr{Construction of multi-relational directed graph} To begin with, we prepare a multi-relational graph representation $\mathcal{G}(\mathcal{V},\mathcal{E},\mathcal{P})$ of the weather stations, \ie, $|\mathcal{V}| = 86$, where we relate them based on two types of relationships, \ie, $|\mathcal{P}| = 2$. First, we connect them based on geographical proximity by inserting an edge between a pair of weather stations if the Euclidean distance between their GPS coordinates is below a threshold. The geographical proximity leads to a symmetric relationship where we acquire $372$ edges. 
Second, we relate them based on the altitude proximity in a similar logic yet we anticipate an asymmetric relationship where the direction of an edge indicates an altitude ascend from one station to another. For both relation types, we adjust the threshold for building connections so that there is not any disconnected node. Consequently, we acquire $1144$ edges for the altitude relationship.
In the experiments, we randomly sample labeled set of nodes, $\mathcal{U}$, from the entire node set, $\mathcal{V}$, with a ratio of $80\%$. Then, we repeat the experiment in this setting for $50$ times in Monte Carlo fashion. The evaluation metrics are then averaged over the series of simulations.


\begin{table}[t]
\begin{minipage}[t]{0.48\linewidth}
\caption{Temperature and Snowfall Prediction Performances}
\label{tbl: temp_snow}
\resizebox{1\textwidth}{!}{
\begin{tabular}{cc|c|c|c}
\toprule
&             & \textbf{RMSE} & \textbf{MAPE} & \textbf{nRMSE}\\
                                                \midrule
{\multirow{2}{*}{\textbf{Temperature}}} & \textsc{LP} & 1.120 & 0.155 & 0.050 \\

 & \textsc{MrP} & 1.040 & 	0.147 & 0.045 \\
 \midrule
{\multirow{2}{*}{\textbf{Snowfall}}} & \textsc{LP} & 194.49 & 0.405 & 0.112 \\

 & \textsc{MrP} & 180.10 & 0.357 & 0.105 \\
 \bottomrule
\end{tabular}}
\end{minipage}
\hspace{0.5cm}
\begin{minipage}[t]{0.42\linewidth}
\caption{Precipitation Prediction Performances}
\label{tbl: prec}
\resizebox{1\textwidth}{!}{
\begin{tabular}{c|c|c|c}
\toprule
            & \textbf{RMSE} & \textbf{MAPE} & \textbf{nRMSE}\\
                                                \midrule
\textsc{LP}-altitude & 381.86 & 0.261 & 0.174 \\
\textsc{LP}-gps & 374.38 & 0.242 & 0.168 \\
\textsc{MrP} & 347.98 & 0.238 & 0.157 \\
 \bottomrule
\end{tabular}}
\end{minipage}
\vspace{-2mm}
\end{table}
\xhdr{Predicting Temperature and Snowfall on Directed Altitude Graph}
We first conduct experiments on a simple scenario where we target predicting temperature and snowfall measurements using altitude relations.
We compare the proposed method to the standard label propagation algorithm, \textsc{LP}, which overlooks asymmetric relational reasoning.
Thus, we aim at evaluating the directional transformation utility of \textsc{MrP} during the neighborhood aggregation, which is mainly gained by the parameter $\tau$ and $\eta$.
We visualize the distribution of measurement changes on the altitude edges in Fig. \ref{fig: Distribution alt}.
Here, the parameter $\tau$ directly corresponds to the mean measurement difference computed along the directed altitude edges since $\eta = 1$. We fit radial basis function (RBF) to the distribution since the residual error in local generative model \eqref{eqn: rel gen model} is assumed to be normal. Then, the parameter $\omega$ is simply associated with the inverse of its variance. We see that the temperature differences in the ascend direction, \ie, $\big\{(x_i - x_j) \: \forall \mathtt{r}(i,j) = \mathtt{altitude\_ascend}\big\}$, has a mean in the negative region. This signifies an expected decrease in temperature values along altitude ascend.

As seen in Table \ref{tbl: temp_snow}, even in the case of single relation type---altitude proximity, incorporating the directionality, \textsc{MrP} manages to enhance predictions over the regression realized by the label propagation, \textsc{LP}. 

\xhdr{Predicting Precipitation on Directed, Multi-relational Graph}
We test \textsc{MrP} in another scenario where we integrate both altitude and geographical proximity relations to predict precipitation measurements on the weather stations.
The prediction performance is compared to the regression by \textsc{LP}, that is accomplished over the altitude relations and GPS relations separately. Since \textsc{MrP} handles both of the relation types and the direction of the edges simultaneously, it achieves a better performance than \textsc{LP}, as seen in Table \ref{tbl: prec}.

\vspace{-4mm}
\subsection{Predicting People's Date of Birth in a Social Network}
\begin{figure}[h]
\centering
\begin{minipage}{0.59\textwidth}%
      \centering
        \includegraphics[width=\linewidth]{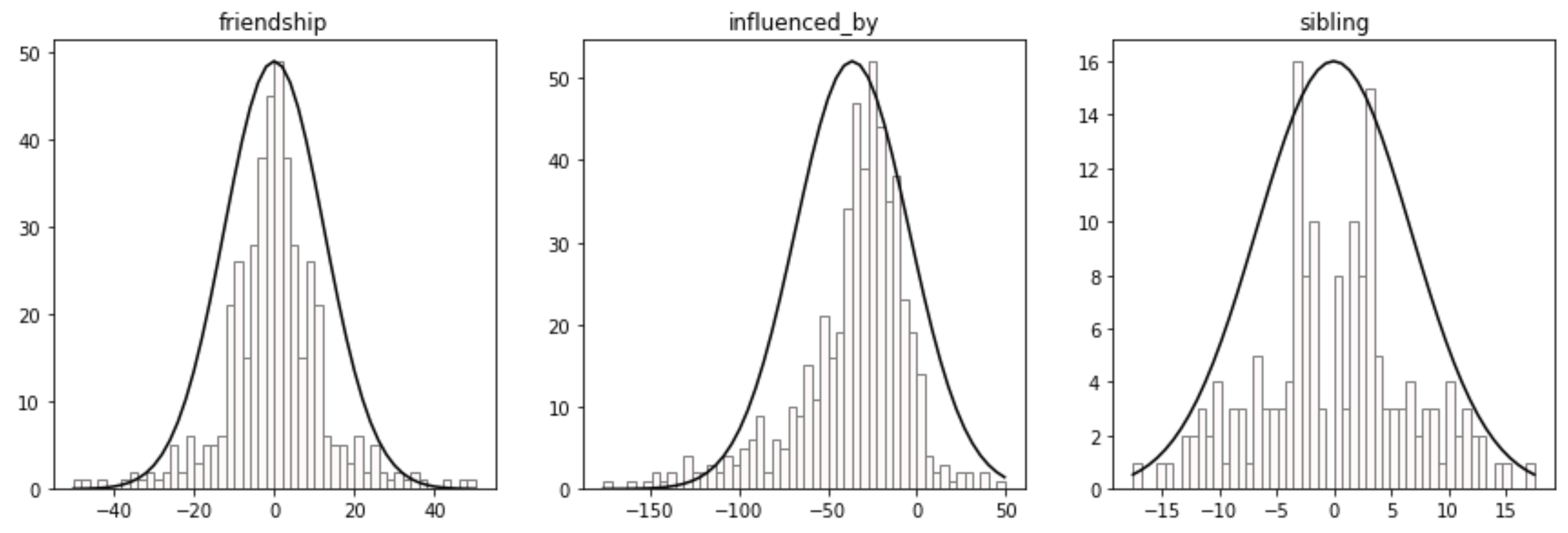}
    \end{minipage}
    \begin{minipage}{0.4\textwidth}   
      \centering
        \includegraphics[width=\linewidth]{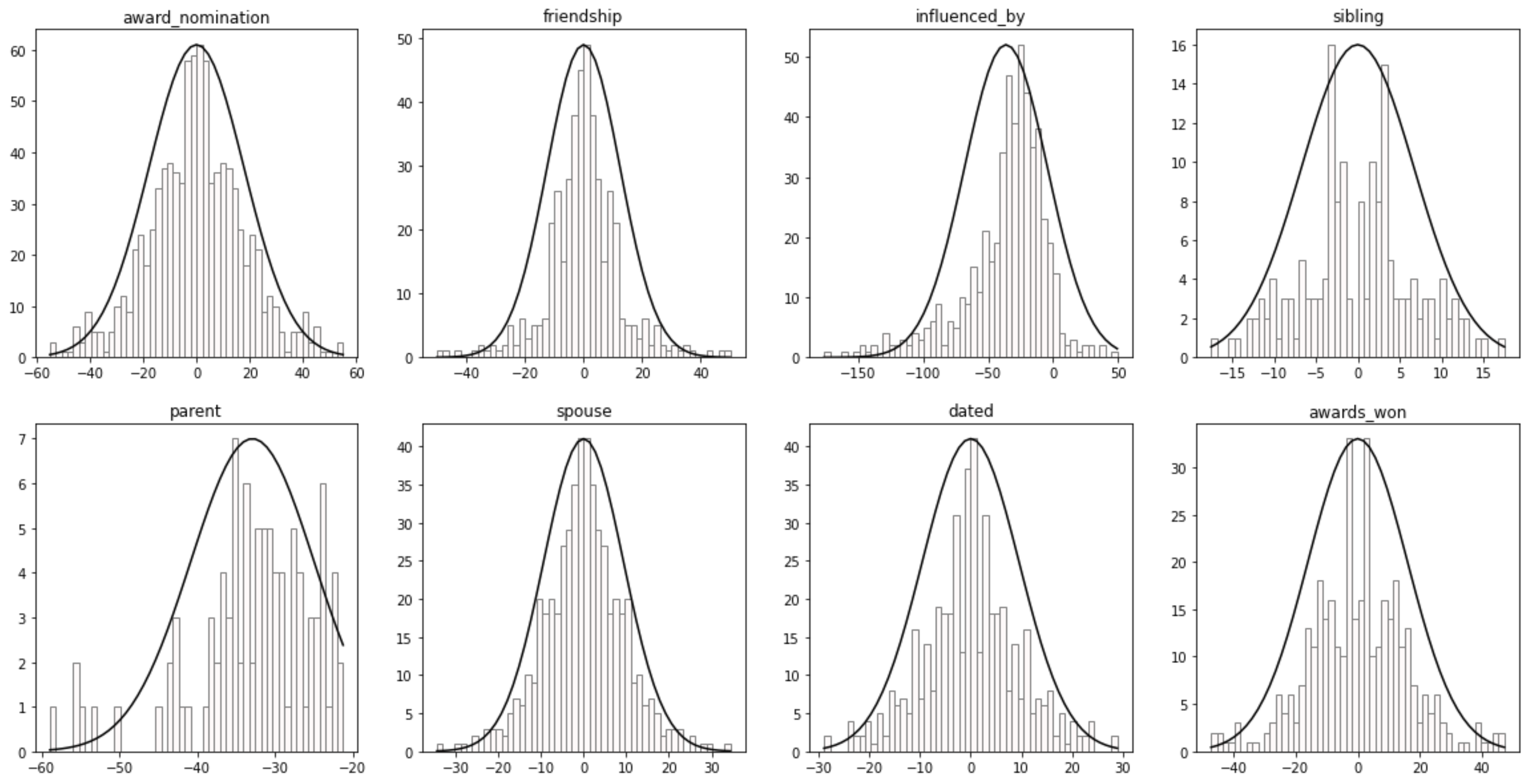}
    \end{minipage}
    \vspace{-2mm}
    \caption{Distribution of `date of birth' difference (year) over different relationships. An RBF is fitted to each.}
\label{fig: date}
\end{figure}

\vspace{-2mm}
We also conduct experiment on a small subset of a relational database called Freebase \cite{toutanova2015observed}. We work on a graph $\mathcal{G}(\mathcal{V},\mathcal{E}, \mathcal{P})$ composing
830 people, \ie, $|\mathcal{V}|= 830$, connected via 8 different types of relationship, \ie, $|\mathcal{P}| = 8$. Here, the task is to predict people's date of birth while it is only known for a subset of people. A fragment of the multi-relational graph is illustrated in Fig. \ref{fig: illust}, where two asymmetric relationships exist: \texttt{influenced\_by} and \texttt{parent}.
Table \ref{tbl: stats} summarizes the statistics for each. For symmetric relationships, the date of birth difference is counted along both directions of an edge, which sets the mean difference to zero. Also, the distributions over certain relation types are visualized in Fig. \ref{fig: date}. 
\begin{table}[t]
\begin{minipage}[t]{0.45\linewidth}
\captionof{table}{Statistics for each relationship, columns respectively: number of edges, mean and variance of `date of birth' difference over the associated relation type.}
\label{tbl: stats}
\centering
\resizebox{1\textwidth}{!}{
\begin{tabular}{llll}
\firsthline
 Relation type                  & edges & mean & variance \\
\midrule                      
\texttt{award\_nomination}       & 454     & 0       &  320.23   \\
\texttt{friendship}             & 221     & 0       &	155.82   \\
\texttt{influenced\_by}          & 528     & -36.25  & 1019.77   \\
\texttt{sibling}                &  83     & 0       &	 45.16   \\
\texttt{parent}                 &  98     & -32.90  &	 62.90   \\
\texttt{spouse}                 & 262     & 0       &	 87.60   \\
\texttt{dated}                  & 231     & 0       &	 90.95   \\
\texttt{awards\_won}             & 183     & 0       &	257.45   \\
\bottomrule
\end{tabular}}
\end{minipage}
\hspace{3mm}
\begin{minipage}[t]{0.55\linewidth}
\captionof{table}{Date of Birth Prediction Performances}
\label{tbl: date}
\centering
\resizebox{1\textwidth}{!}{
\begin{tabular}{rr|c|c|c}
\firsthline
        & Relation type   & \textbf{RMSE} & \textbf{MAPE} & \textbf{nRMSE}\\
                                                \midrule
 & \texttt{award\_nomination}    & 32.43 &  0.011    & 0.115  \\
 & \texttt{friendship}           & 31.92 &  0.011    & 0.113  \\
 & \texttt{influenced\_by}       & 30.29 &  0.012    & 0.108  \\
 & \texttt{sibling}              & 32.69 &  0.012    & 0.116  \\
\textsc{LP} & \texttt{parent}    & 33.62 &  0.013    & 0.119  \\
 &  \texttt{spouse}               & 31.45 &  0.011    & 0.112  \\
 &  \texttt{dated}                & 31.70 &  0.011    & 0.113  \\
 &  \texttt{awards\_won}          & 33.04 &  0.012    & 0.117  \\
 &  union                         & 24.22 &  0.008    & 0.086  \\
\midrule
\textsc{MrP}&  & 15.62 & 0.005 & 0.055 \\
 \bottomrule
\end{tabular}}
\end{minipage}
\vspace{-4mm}
\end{table}

In the experiments, we randomly select the set of people whose date of birth is initially known, $\mathcal{U}$, with a ratio of $50\%$ in $\mathcal{V}$. We again report the evaluation metrics that are averaged over a series of experiments repeated for 50 times.
We compare performance of \textsc{MrP} to the regression of date of birth values obtained with label propagation \textsc{LP}. We run \textsc{LP} over the edges of each relation type separately and also at the union of those. The results are given in Table \ref{tbl: date}. Based on the results, we can say that the most successful relation types for predicting the date of birth seems to be \texttt{influenced\_by} and \texttt{spouse} using \textsc{LP}. Nonetheless, when \textsc{LP} operates on the union of the edges provided by different type of relationships, it performs better than any single type. \textsc{MrP} is able to surpass this record by enabling a relational neighborhood aggregation over different types of edges. Once again, we argue that its success is due to the fact that it regards asymmetric relationships, here encountered as \texttt{influenced\_by} and \texttt{parent}. In addition, it assigns different level of importance to the predictions collected through different types of relationships based on the uncertainty estimated over the observed data.
\vspace{-2mm}
\vspace{-10mm}
\section{Conclusion}
\vspace{-2mm}
In this study, we proposed \textsc{MrP}, a propagation algorithm working on multi-relational and directed graphs for regression of continuous node features and we show its superior performance on multi-relational data compared to standard propagation algorithm. 
It is possible to generalize the proposed approach for node embedding learning and then for the node classification tasks. The augmentation of the computational graph of the propagation algorithm using multiple types of directed relationships provided by the domain knowledge permits anisotropic operations on graph, which is claimed to be promising for future directions in graph representation learning \cite{dwivedi2020benchmarking}.

\bibliographystyle{splncs03}
\vspace{-2mm}
{\small
\bibliography{refs.bib}
}

\section*{Supplementary Material}

\subsection*{Negative Log-likelihood Estimation with Simple Local Generative Model}
The simple local generative model is defined as $x_i = x_j + \epsilon$
where $(i,j) \in \mathcal{E}$. Since the additive noise is Gaussian, \ie, $\epsilon \sim \mathcal{N}(0, \sigma^2_{ij})$, we express the likelihood of a node's feature given its neighbor's as
\begin{equation}
    \mathrm{p}(x_j|x_i) = \frac{1}{\sqrt{2 \pi \sigma_{ij}^2}}
    \mathrm{exp}\Big(-\frac{(x_j-x_i)^2}{2\sigma^2_{ij}}\Big).
\end{equation}
Once this is plugged in the negative log-likelihood minimization problem \eqref{eqn: MLE}, we obtain
\begin{align}
\operatorname*{argmin}_{x_i} \quad
\sum_{(i,j) \in \mathcal{E}} \Big(
-\mathrm{log}\big(\frac{1}{\sqrt{2 \pi \sigma_{ij}^2}}\big) + \frac{(x_j-x_i)^2}{2\sigma^2_{ij}} \Big).
\end{align}
The first term inside the sum is inherently omitted since it does not depend on the variable we minimize, which results in Problem \eqref{eqn: minim}.

\subsection*{The Likelihoods leading to the First-order Relational Bayesian Estimate}
In the relational local generative model \eqref{eqn: rel gen model}, the additive noise is Gaussian. Thus, one can express the likelihood of a relational neighbor as follows:
\begin{equation}
\label{eqn: likelihood}
    \mathrm{p}(x_j|x_i) =
    \Bigg\{
      \begin{array}{l}
      \sqrt{\frac{\omega_{\mathtt{p}}}{2\pi}} \:
      \mathrm{exp}\Bigg( -\cfrac{\omega_\mathtt{p}}{2}\Big(x_i - \eta_\mathtt{p}x_j -\tau_\mathtt{p}\Big)^2\Bigg)
       \quad	 \text{if  } \mathtt{r}(i,j) = \mathtt{p}\\ 
        \sqrt{\frac{\omega_{\mathtt{p}}\eta_\mathtt{p}^2}{2\pi}}\:
      \mathrm{exp}\Bigg( -\cfrac{\omega_\mathtt{p}\eta_\mathtt{p}^2}{2}\Big(x_i - \cfrac{x_j}{\eta_\mathtt{p}} +\cfrac{\tau_\mathtt{p}}{\eta_\mathtt{p}}\Big)^2
      \Bigg)
       \quad	 \text{if  } \mathtt{r}(i,j) = \mathtt{p}^{-1},
      \end{array}
\end{equation}
where we apply the change of parameter: $\omega_p = 1/\sigma^2_{\mathtt{p}}$.

The first-order Bayesian estimate of node's feature can be found by minimizing the negative log-likelihood as in \eqref{eqn: MLE}. For this purpose, the likelihoods in \eqref{eqn: likelihood} are substituted, which results in the objective \eqref{eqn: rel Bayes problem}.

\subsection*{Likelihood of Neighboring Node Features given Relational Parameters}
We express the likelihood of a pair of values $(x_i, x_j)$ belonging to the nodes connected by relation type $p$ using the model \eqref{eqn: rel gen model} as
\begin{align}
\label{eqn: param likelihood}
    \mathrm{p} \Big(
    (x_i,x_j) \: | \: \mathtt{r}(i,j)=\mathtt{p}
    \: \big| \: \tau_\mathtt{p}, \eta_\mathtt{p}\, \omega_\mathtt{p}
    \Big) =
    \sqrt{\frac{\omega_{\mathtt{p}}}{2\pi}} \:
      \mathrm{exp}\Bigg( -\cfrac{\omega_\mathtt{p}}{2}\Big(x_i - \eta_\mathtt{p}x_j -\tau_\mathtt{p}\Big)^2\Bigg).
\end{align}

\subsection*{Derivation of Relational Parameters}
Maximum likelihood estimation of model parameters belonging to relation type $\mathtt{p} \in \mathcal{P}$ is realized over the node pairs connected by that relationship as stated in Problem \eqref{eqn: rel MLE}. We rewrite the problem as minimization of negative log-likelihood and encapsulate the objective as follows: 
\begin{align}
\label{eqn: sup rel}
    \min_{\tau_\mathtt{p}, \eta_\mathtt{p}\, \omega_\mathtt{p}} \quad
    \sum_{i,j \in \mathcal{V}\: | \: \mathtt{r}(i,j)=\mathtt{p}}
    \mathcal{L}_{ij}(\tau_\mathtt{p}, \eta_\mathtt{p}\, \omega_\mathtt{p})
\end{align}
where the loss inside the sum is defined for a pair of neighboring nodes as
$$\mathcal{L}_{ij}(\tau_\mathtt{p}, \eta_\mathtt{p}\, \omega_\mathtt{p}) = -\mathrm{log}\Big(\mathrm{p} \Big(
    (x_i,x_j)
    \: \big| \: \tau_\mathtt{p}, \eta_\mathtt{p}\, \omega_\mathtt{p}
    \Big)\Big).
$$
Plugging the likelihoods \eqref{eqn: param likelihood} in the loss above, the solution of the problem in \eqref{eqn: sup rel} can be found by setting the gradient of the sum of losses over the node pairs connected by the relation type $\mathtt{p}$ to zero:
\begin{align}
    \cfrac{\sum_{\mathtt{r}(i,j)=\mathtt{p}}\partial\mathcal{L}_{ij}}{\partial \tau_\mathtt{p}} = & \quad
     \sum_{\mathtt{r}(i,j)=\mathtt{p}}
     -\omega_\mathtt{p} (x_i - \eta_\mathtt{p} x_j - \tau_p)
    &= 0,\\
    \cfrac{\sum_{\mathtt{r}(i,j)=\mathtt{p}} \partial \mathcal{L}_{ij}}{\partial \eta_\mathtt{p}} = & \quad \sum_{\mathtt{r}(i,j)=\mathtt{p}}
    -\omega_\mathtt{p} x_j (x_i - \eta_\mathtt{p} x_j - \tau_p)
    &= 0,\\
    \cfrac{\sum_{\mathtt{r}(i,j)=\mathtt{p}}\partial\mathcal{L}_{ij}}{\partial \omega_\mathtt{p}} = & \quad
    \sum_{\mathtt{r}(i,j)=\mathtt{p}}
    -\cfrac{1}{2\omega_\mathtt{p}} + \frac{1}{2}(x_i - \eta_\mathtt{p} x_j - \tau_p)^2
    &= 0.
\end{align}
Consequently, the set of parameters $\{\tau_\mathtt{p}, \eta_\mathtt{p}\, \omega_\mathtt{p}\}$ associated with relation $\mathtt{p}$ are solved as equivalent to the parameters of a linear regression problem, which results in \eqref{eqn:eta}.

\end{document}